\documentclass[11pt]{article}

\usepackage{graphicx}
\usepackage{amssymb}
\usepackage{amsmath}

\usepackage{lineno}
\usepackage{optidef}
\usepackage{adjustbox}
\usepackage{subcaption}
\usepackage{natbib}
\usepackage[superscript]{cite}
\usepackage{tabu}
\usepackage[flushleft]{threeparttable}
\usepackage{booktabs}
\usepackage{longtable}
\usepackage{multirow}
\usepackage{float}
\restylefloat{table}
\usepackage{geometry}
\usepackage{setspace}
\usepackage{adjustbox}
\usepackage{lscape}
\usepackage{bm}
\usepackage{xcolor}
\usepackage{authblk}

\date{ }
\begin{document}


\title{Hybrid Density- and Partition-based Clustering Algorithm for Data with Mixed-type Variables}


\author[1,2]{Shu Wang}
\author[3,4,5]{Jonathan G. Yabes}
\author[3,4,5]{Chung-Chou H. Chang}
\affil[1]{\small{Department of Biostatistics, College of Public Health and Health Professions, University of Florida}}
\affil[2]{\small{University of Florida Health Cancer Center}}
\affil[3]{\small{Department of Biostatistics, Graduate School of Public Health, University of Pittsburgh}}
\affil[4]{\small{Department of Medicine, School of Medicine, University of Pittsburgh}}
\affil[5]{\small{Department of Clinical and Translational Science, School of Medicine, University of Pittsburgh}}

\renewcommand\Authands{ and }

\maketitle
\renewcommand{\abstractname}{\vspace{-\baselineskip}}

\begin{abstract}
Clustering is an essential technique for discovering patterns in data. The steady increase in amount and complexity of data over the years led to improvements and development of new clustering algorithms. However, algorithms that can cluster data with mixed variable types (continuous and categorical) remain limited, despite the abundance of data with mixed types particularly in the medical field. Among existing methods for mixed data, some posit unverifiable distributional assumptions or that the contributions of different variable types are not well balanced.

We propose a two-step hybrid density- and partition-based algorithm (HyDaP) that can detect clusters after variables selection. The first step involves both density-based and partition-based algorithms to identify the data structure formed by continuous variables and recognize the important variables for clustering; the second step involves partition-based algorithm together with a novel dissimilarity measure we designed for mixed data to obtain clustering results. Simulations across various scenarios and data structures were conducted to examine the performance of the HyDaP algorithm compared to commonly used methods. We also applied the HyDaP algorithm on electronic health records to identify sepsis phenotypes.\\
\textit{KEY WORDS:} Clustering, Mixed data, Variable selection
\end{abstract}


\section{Introduction}
\label{S:1}

In precision medicine, the prevention and treatment strategies are tailored according to individual characteristics. Such practice has been greatly improved by using information obtained from large databases \citep{national2011toward} including electronic health record (EHR) which contains patient information such as demographics, daily charts, medical history, lab results, medication use, billing information and others \citep{hayrinen2008definition}. In order to efficiently process data and extract useful information, machine learning methods are often applied \citep{coorevits2013electronic}. Clustering is an important aspect of unsupervised machine learning methods which aims to uncover hidden patient subgroups that may have different diagnoses and treatment responses in EHR data. Further investigations on these subgroups together with current clinical guidelines could help design precision medicine strategies to further assist physicians in providing better patient care \citep{jensen2012mining}.

The basic concept of clustering is to divide individuals into a number of subgroups such that individuals within the same subgroup have more similar characteristics, as defined by a set of variables, than the individuals who belong to different subgroups. One of the main challenges in clustering is how to define ``dissimilarity'' between subjects with data of mixed variable types (continuous and categorical). If all variables are continuous, we can view the collection of information from an individual as a data point, or a vector of variables in a high-dimensional covariate space. The \textit{distance} between the data points of two individuals is used to determine the \textit{dissimilarity} between these two subjects so that a closer distance indicates lower dissimilarity. If all variables are categorical, \textit{dissimilarity measures} (or \textit{similarity measures}) were proposed to evaluate how often two individuals are in the same category among those variables. In this context we will use ``distance" and ``dissimilarity" interchangeably. Gower distance \citep{gower1971general}, distance defined in factorial analysis of mixed data (FAMD) \citep{pages2014multiple}, and K-prototypes \citep{huang1998extensions} are possible methods to address the above mentioned issue.

Gower distance was proposed to measure dissimilarity between subjects with mixed types of variables. The distance measure used in FAMD can be applied on mixed data as well, even though FAMD was not originally intended for clustering. Distance measure defined in K-prototypes is similar to Gower distance, but it incorporates a user-defined weight for each type of variables. Therefore, K-prototypes assumes that all categorical variables have the same weight, and that all continuous variables have the same weight. This design may not be practical if within the same variable type, some are clinically more important than others in terms of clustering.

Finite mixture model (FMM) \citep{mccutcheon1987latent, moustaki1996latent} is a model-based clustering method that bypasses the challenge of defining dissimilarity between subjects with mixed types of variables. It assumes that the data is a mixture of several parametric distributions. The unknown distributional parameters including cluster membership can be solved via maximizing likelihood using the expectation-maximization (EM) algorithm. Moreover, it is able to transfer the task of selecting the optimal number of clusters into model selection problem which is much more straightforward. However, its main drawback is that all the distributional assumptions are conditional on the unknown cluster, making those assumptions unverifiable from the data.

In order to identify cluster memberships, it is also important to know the underlying data structure. For example, whether distinct clusters exist in the feature space; or if no natural clusters exist but the data is heterogeneous enough to be partitioned. Such information is crucial in understanding data, selecting clustering methods, and interpreting clustering results. However, to our knowledge, none of the existing methods incorporates this data structure information into clustering.

To address the limitations of the existing methods, we propose a Hybrid Density- and Partition-based (HyDaP) algorithm to identify clusters for data with mixed types of variables and use this method to discover sepsis phenotypes using demographic and clinical data in EHR for sepsis patients at university affiliated hospitals.

In Section~\ref{sec:review} we introduce the most commonly used dissimilarity measures and clustering algorithms; in Section~\ref{sec:hydap} we define three data structures and propose a new clustering algorithm, HyDaP; in Section~\ref{sec:sim} we present performance comparisons among different methods under various simulation settings; in Section~\ref{sec:real} we demonstrate the use of HyDaP algorithm to identify sepsis phenotypes; and Section~\ref{sec:dis} is discussion.

\section{Review of dissimilarity measures and clustering algorithms}
\label{sec:review}

In this section, we briefly review some existing dissimilarity measures and clustering algorithms. In addition, we discuss the pros and cons of each measure or algorithm.

\subsection{Dissimilarity measures}

Minkowski distance is a family of dissimilarity measures for numeric variables. Let $\mathbf{x}_i$ be a vector $(x_{i1}, x_{i2}, ..., x_{ip})^T$ representing $p$ variables of subject $i$. For subjects $i$ and $ i^{\prime}$, Minkowski distance between the two is defined as follows:
$$d(\mathbf{x}_i,\mathbf{x}_{ i^{\prime}})=\left(\sum_{j=1}^{p}|x_{ij}-x_{ i^{\prime}j}|^m\right)^{\frac{1}{m}}, m\geqslant1$$
where $m$ is related to the \textit{shape of unit circle} which is a two-dimensional contour with every point on the contour at distance of 1 from the center $(0, 0)$. Different choices of $m$ lead to different distance measures. For example, $m=2$ leads to the famous Euclidean distance which is intuitive and able to represent physical distances. When $m=1$, we obtain Manhattan distance which is often used to detect hyperrectangular clusters. When $m\to\infty$, we obtain Chebyshev (maximum) distance which is the same as chess board distance since it is defined as the greatest value of the differences among all dimensions. A potential problem of using the Minkowski distance is that variables with larger variances tend to dominate the others \citep{xu2005survey,shirkhorshidi2015comparison}, therefore, it is recommended to perform variable standardization (that is, rescale the variable by dividing by its standard deviation) before applying this measure.

Other dissimilarity measures for numeric variables include cosine similarity measure, Pearson correlation, Mahalanobis distance, to name a few. Cosine similarity measures the angle between two vectors regardless of vector magnitudes. It is usually applied if we are not interested in magnitudes, for example, for text mining as it captures text meanings instead of counting numbers \citep{xu2005survey, han2011data}. Pearson correlation is usually used in clustering gene expression data \citep{xu2005survey}, but it is sensitive to outliers. Mahalanobis distance is scale-invariant, and takes into account variable correlations.

When variables are all categorical, simple matching dissimilarity is usually used:
$d(\mathbf{x}_i,\mathbf{x}_{ i^{\prime}})=\sum_{j=1}^{p}\delta(x_{ij},x_{ i^{\prime}j}),$ where $\delta(x_{ij},x_{ i^{\prime}j})= I(x_{ij}\ne x_{ i^{\prime}j})$ indicating whether variable $j$ are the same for individuals $i$ and $ i^{\prime}$.

None of above-mentioned dissimilarity measures can be applied to mixed data. Gower distance was proposed to calculate the distance between subjects with mixed types of variables. Let $\mathbf{X}$ be a data matrix with $n\times p$ dimensions. Let the first $h$ variables of $\mathbf{X}$ be continuous and the $(h+1)^{th}$ to $p^{th}$ variables be multilevel categorical variables or symmetric binary variables. Let $\mathbf{X}_j$ be a vector $(x_{1j}, x_{2j}, ..., x_{nj})^T$ representing variable $j$. Gower distance between individuals $i$ and $i^{\prime}$ is defined as:
$$d(\mathbf{x}_i,\mathbf{x}_{ i^{\prime}})= \sum_{j=1}^{p}{d_j(\mathbf{x}_i,\mathbf{x}_{ i^{\prime}})},$$
where
\[ d_j(\mathbf{x}_i,\mathbf{x}_{ i^{\prime}}) =
\begin{cases}
\frac{|x_{ij}-x_{ i^{\prime}j}|}{max(\mathbf{X}_j)-min(\mathbf{X}_j)} & \quad \text{if } j \in \{1, 2, ..., h\}\\
I(x_{ij} \neq x_{ i^{\prime}j}) & \quad \text{if } j \in \{h+1, h+2, ..., p\},
\end{cases}
\]

$$max(\mathbf{X}_j)=x_{i^\star j} \text{ if } x_{i^\star j}\geqslant x_{ij} \text{ for all } i,$$
$$min(\mathbf{X}_j)=x_{i^\star j} \text{ if } x_{i^\star j}\leqslant x_{ij} \text{ for all } i.$$

Gower distance for an \textit{asymmetric} binary variable is calculated differently. Asymmetry occurs when similarity within one level is perceived to be higher compared to the other level. For example, breast cancer (yes/no) could be viewed as an asymmetric binary variable since individuals with breast cancer are much more similar than those without breast cancer (which could include men and women, adolescents and elder people). If variable $j$ is an asymmetric binary variable, then the Gower distance between individuals $i$ and $ i^{\prime}$ with respect to this variable is defined as:
\[ d_j(\mathbf{x}_i,\mathbf{x}_{ i^{\prime}}) =
\begin{cases}
0 & \quad \text{if } x_{ij} = x_{ i^{\prime}j} \text{ and they are the level with larger similarity}\\
1 & \quad \text{otherwise}.
\end{cases}
\]
In practice, there is one issue in applying Gower distance: as we will later show in simulations, Gower distance tends to give much larger weights to categorical variables than to continuous ones. This is because the distance due to a categorical variable is always 0 or 1, the minimum and the maximum of possible distance values, granting categorical variables more power in distinguishing subjects.

Another distance that could be used for mixed types of variables is the distance defined in FAMD:
$$d^2(\mathbf{x}_i,\mathbf{x}_{ i^{\prime}})= \sum_{j=1}^{p}{d_j^2(\mathbf{x}_i,\mathbf{x}_{ i^{\prime}})},$$
where
\[ d_j^2(\mathbf{x}_i,\mathbf{x}_{ i^{\prime}}) =
\begin{cases}
(x_{ij}-x_{i^{\prime}j})^2 & \quad \text{if } j \in \{1, 2, ..., h\}\\
\sum_{l=1}^{C_j}\frac{1}{p_{jl}}(\frac{y_{ijl}}{p_{jl}}-\frac{y_{i^{\prime}jl}}{p_{jl}})^2 & \quad \text{if } j \in \{h+1, h+2, ..., p\},
\end{cases}
\]
$$y_{ijl} = I(x_{ij}=L_{jl})\text{, } \sum_{l=1}^{C_j}y_{ijl} = 1\text{; } j \in \{h+1, h+2, ..., p\}$$
$C_j$ is number of levels of categorical variable $j$; $p_{jl}$ is proportion of $l^{th}$ category of variable $j$; $L_{jl}$ is $l^{th}$ category of variable $j$.

\subsection{K-means-based clustering algorithms}

K-means \citep{macqueen1967some} is the most well-known and applied clustering method in practice. The basic idea is to partition subjects with respect to minimizing the within-cluster sum of squares (WCSS). This algorithm is very efficient and has been the root of many later developed ones. It is usually used together with Euclidean distance. To cluster categorical data, K-modes \citep{huang1998extensions} algorithm was developed by replacing Euclidean distance with simple matching dissimilarity measure, and replacing mean with mode to represent cluster centers.

To identify clusters with mixed types of variables, the partition around medoids (PAM) \citep{kaufman2009finding} has been proposed. PAM is a modification of K-means with a different definition of cluster centers.
Unlike K-means which uses within-cluster mean to represent its centers, PAM uses \textit{medoids} which are actual data points in the dataset. This makes defining centers of categorical variables possible.
Moreover, medoids are analogous to medians and hence PAM is more robust to outliers. One drawback however is that PAM is computationally intensive and inefficient, making it less ideal for processing large data sets.

K-prototypes algorithm is another modified version of K-means with the ability of handling mixed types of variables. Its centers are called \textit{prototypes}, which use within-cluster mean to represent continuous variables and mode for categorical variables. The distance between subjects $i$ and $i^{\prime}$ is defined as:
$$d(\mathbf{x}_i,\mathbf{x}_{ i^{\prime}})= \sum_{j=1}^{h}{d_j(\mathbf{x}_i,\mathbf{x}_{ i^{\prime}})}+\gamma\sum_{j=h+1}^{p}{d_j(\mathbf{x}_i,\mathbf{x}_{ i^{\prime}})},$$
where
\[ d_j(\mathbf{x}_i,\mathbf{x}_{ i^{\prime}}) =
\begin{cases}
(x_{ij}-x_{ i^{\prime}j})^2 & \quad \text{if } j \in \{1, 2, ..., h\}\\
I(x_{ij} \neq x_{ i^{\prime}j}) & \quad \text{if } j \in \{h+1, h+2, ..., p\},
\end{cases}
\]
and $\gamma$ is a user-defined weight parameter for categorical variables. K-prototypes lacks flexibility in variables weights as it assumes equal importance for variables of the same type. Moreover, the tuning parameter $\gamma$ is user-defined rather than data-driven.

\subsection{Hierarchical clustering}

Hierarchical clustering is another category of clustering methods. It first grows a dendrogram which is a tree-like diagram showing hierarchical structure of subjects and then cuts the dendrogram to obtain clusters. One advantage of hierarchical clustering is that the generated dendrogram is very informative and provides information of cluster structure besides cluster assignments. Its disadvantages include no global objective function, a greedy type of procedure, the sensitivity to outliers, and inefficient for large data sets.

\subsection{Extended clustering framework}

In many situations researchers are also often interested in variables' importance, not just cluster identification. Motivated by this interest, sparse clustering framework \citep{witten2010framework} was proposed. It incorporates feature selection through a Lasso-type penalty, and adds variable weights to the objective function:
\begin{maxi*}
{\mathbf{w};\mathbf{\Theta} \in D}{\sum_{j=1}^{p}w_j f_j(\mathbf{X}_j;\mathbf{\Theta})}
{}{}
\addConstraint{\|\mathbf{w}\|^2\leqslant1,\|\mathbf{w}\|_1\leqslant s, w_j\geqslant0 \textit{ } \forall j},
\end{maxi*}
where $n$ is number of subjects; $p$ is number of features; $\mathbf{w}=(w_1,w_2,...,w_p)^T$ is the weight vector; $\mathbf{\Theta}$ is a parameter vector restricted to lie in a set $D$; $f_j(\mathbf{X}_j;\mathbf{\Theta})$ is some function that involves feature $j$ only; and $s$ is a $L1$ norm restriction, which is a tuning parameter in the algorithm. We could plug in many algorithms like K-means, hierarchical clustering into this framework to obtain sparse version algorithms. One of the main attractions of sparse clustering is that it conducts data-driven variable selection and clustering simultaneously. However, the selection of tuning parameter $s$ may not be straightforward.

Many partition-based algorithms require pre-specification of $K$, the optimal number of clusters, but how to choose it is another important question. The consensus clustering framework \citep{monti2003consensus, wilkerson2010consensusclusterplus} can help determine number of clusters and obtain cluster memberships simultaneously. In addition, it can assess stability of discovered clusters. Consensus clustering incorporates results from multiple runs of an inner-loop clustering algorithm (e.g., K-means) on sub-sampled subjects. For each pair of subjects, a consensus index is obtained by calculating proportion of times the pair was assigned to the same cluster among times both pair members were sampled. The consensus index can then serve as a similarity measure and subjected to a hierarchical clustering algorithm to form final clusters. Choosing $K$ is achieved by checking the consensus matrix heatmaps and cluster-consensus values. The number of clusters that yields the cleanest heatmap and highest cluster-consensus values is preferred.

\subsection{Density-based clustering}

Another important category of clustering methods is density-based clustering. All above-mentioned algorithms are distance-based methods which are more appropriate for detecting clusters that are convex shaped and with similar sizes and densities. If the underlying clusters have arbitrary shapes, density-based clustering algorithms may work better. Density-based spatial clustering of applications with noise (DBSCAN) \citep{ester1996density} and ordering points to identify the clustering structure (OPTICS) \citep{ankerst1999optics} are two widely used density-based algorithms. DBSCAN does not need input of $K$, and it is robust to noise. However, it is not well suited for high dimensional data or for clusters with varying densities. OPTICS is an improved method which can detect clusters with varying densities while not being over sensitive to its user-specified tuning parameters.

\subsection{Model-based clustering}

FMM is a model-based clustering method assuming that the data is consists of $K$ latent clusters. Its density function is defined as: $$f(\mathbf{X})=\sum_{k=1}^{K}\pi_kg_k(\mathbf{X}),$$
where $\pi_k$ is the cluster mixture probability, $\sum_{k=1}^{K}\pi_k=1$; and $g_k$ is the conditional distribution given cluster $k$.
For a sample of size $n$, the log-likelihood can be written as:
$$L=\sum_{i=1}^{n}\log f(\mathbf{x}_i)=\sum_{i=1}^{n}\log\sum_{k=1}^{K}\pi_kg_k(\mathbf{x}_i).$$
FMM assumes conditional independence given cluster $k$, that is, $g_k(\mathbf{x}_i)=\prod_{j=1}^{p}g_k(x_{ij})$, and the
EM algorithm is usually used to obtain the MLE. The posterior probability of each subject belonging to each cluster can be calculated as: $$\hat{p}(k|\mathbf{x}_i)=\frac{\hat{\pi_k}\hat{g_k}(\mathbf{x}_i)}{\sum_{k=1}^{K}\hat{\pi_k}\hat{g_k}(\mathbf{x}_i)}.$$
Subjects are then assigned to the cluster with which the posterior probability is the largest. These probabilities help discriminate core subjects (those with high probability of belonging to assigned cluster) and border subjects (those with low probability of belonging to assigned cluster) within each cluster. Given the parametric form of FMM, formal inference is possible. In addition, selecting the number of clusters becomes a model selection problem. The main drawback however is its unverifiable distributional assumptions; all the inferences are conducted conditional on unknown cluster assignments.

There are some other approaches to handle mixed types of variables. These include categorizing all continuous variables \citep{haripriya2015integrating} or converting categorical variables into continuous or dummy variables and then treat the dummy variables as continuous \citep{hennig2013find}. However, both ideas will lead to information loss.
Another common idea is to cluster continuous part of the data and categorical part separately. The final clusters are obtained by ensembling these two sets of clustering results \citep{reddy2012clustering}. This method impractically weigh continuous and categorical variables equally and ignore possible mutual influences between the two variable types.

\section{Proposed hybrid density- and partition-based clustering (HyDaP) algorithm for mixed data}
\label{sec:hydap}

To address the limitations of the existing clustering methods in handling data containing mixed types of variables, we propose a hybrid density- and partition-based clustering (HyDaP) algorithm which consists of a \textbf{pre-processing step} (step 1) and a \textbf{clustering step} (step 2). The pre-processing step identifies the data structure formed by continuous variables and recognizes the important variables for clustering. In the clustering step, our proposed dissimilarity measure is used to obtain a dissimilarity matrix, which can be fed into PAM to obtain the final results. We describe the HyDaP algorithm in detail below.

\subsection{Pre-processing step (Step 1)}
\begin{figure}[H]
     \centering
     \includegraphics[width=0.9\textwidth]{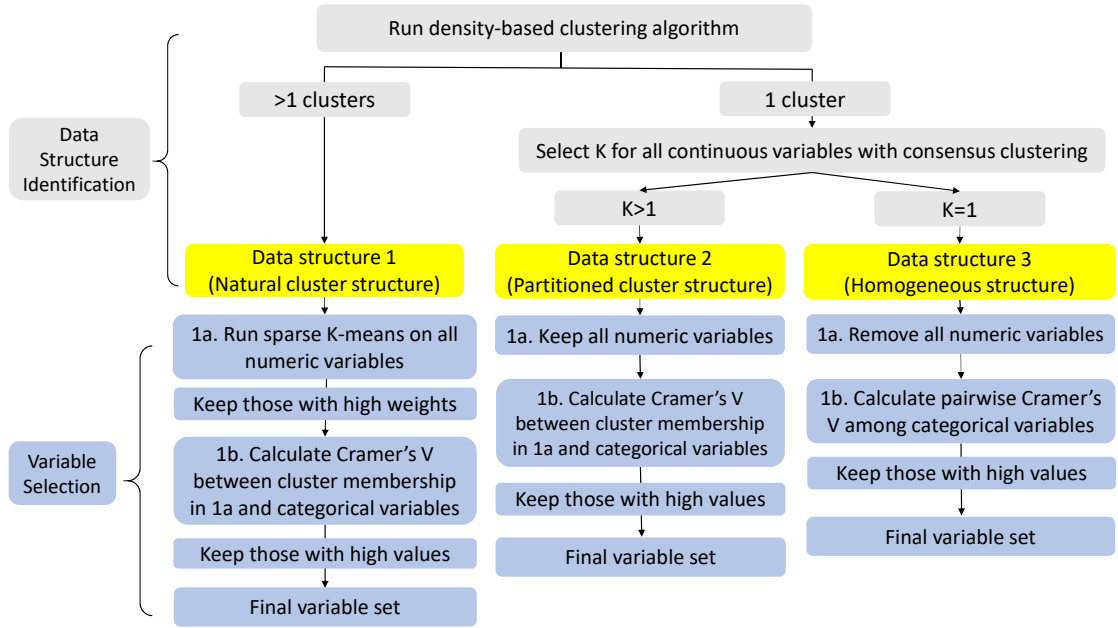}
     \caption{Flowchart of Step 1 of the HyDaP algorithm}
     \label{fig:flow}
\end{figure}

To help with variable selection and better understand the data set, we first define 3 data structures for the space spanned by the continuous variables as: \textit{natural cluster structure} (data structure 1); \textit{partitioned cluster structure} (data structure 2); and \textit{homogeneous structure} (data structure 3). Once the data structure is known, we apply tailored variable selection procedures. At the end of the pre-processing step, a set of selected variables will proceed to the clustering step (step 2) (Figure~\ref{fig:flow}).

\subsubsection{Data structure identification}
\begin{figure}[H]
     \centering
     \includegraphics[width=0.9\textwidth]{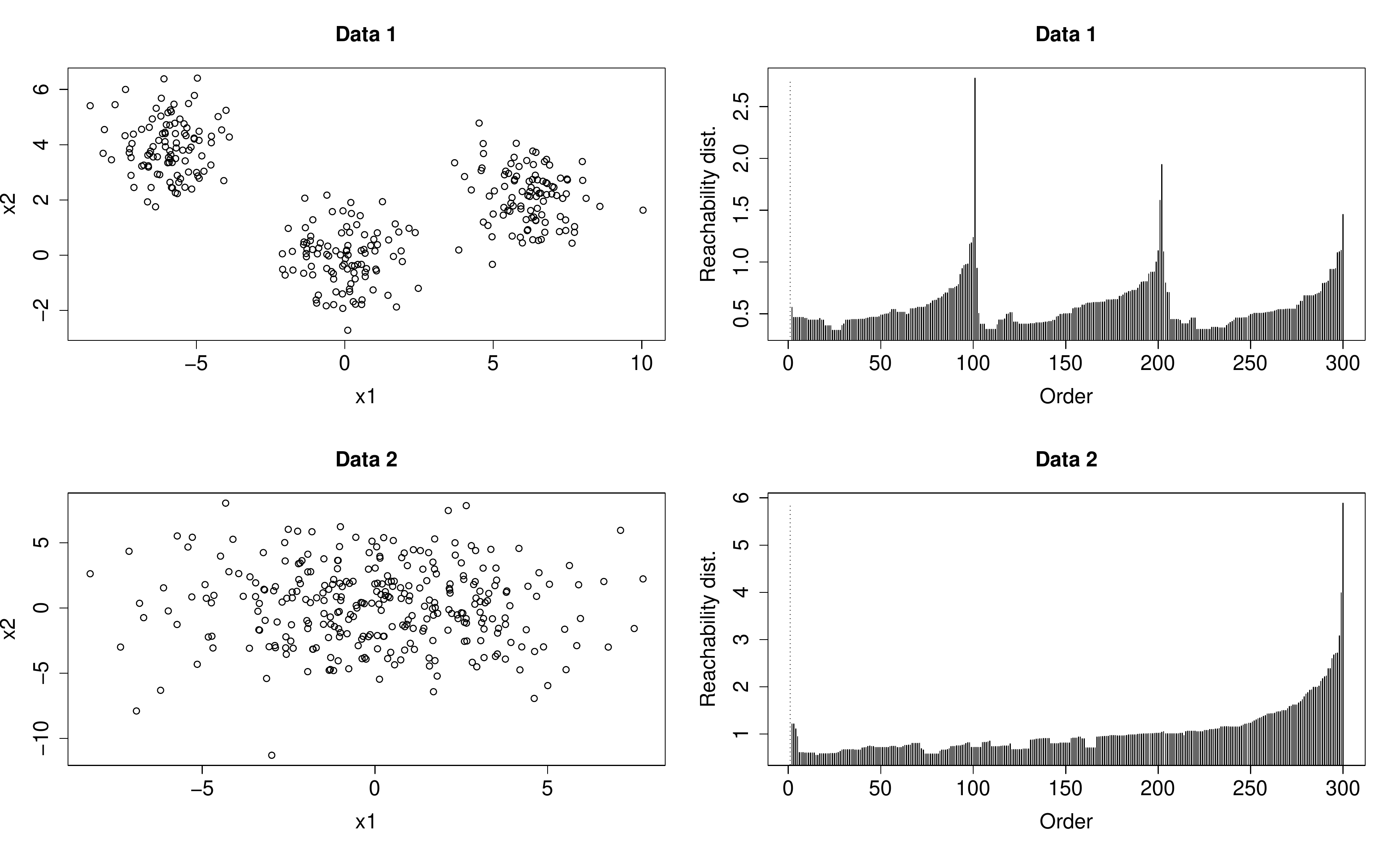}
     \caption{Illustration of different reachability plots}
     \label{data12}
\end{figure}

Data spanned in the covariate space of continuous variables can be divided into two scenarios: with and without natural clusters. A hypothetical example of these two scenarios is depicted in Figure~\ref{data12}. We can observe that both Data 1 and Data 2 contain two variables, but natural clusters only exist in Data 1. Although this conclusion is straightforward for Data 1 and Data 2, when data is spanned in a high-dimensional space, it is impossible to visually examine existence of natural clusters.
Therefore, we use a density-based clustering algorithm (e.g., OPTICS) and resulted reachability plot to help understand the spatial structure of the data. Reachability plot is a bar plot showing ordered reachability distances among subjects \citep{ankerst1999optics}. A reachability plot provides an overall 2-dimensional spatial structure of a dataset regardless of its original dimensions. The horizontal axis of the plot is the processing order and the vertical axis is the reachability distance. Each trough on the reachability plot can be viewed as a single cluster. Edges between two side-by-side troughs represent the distance between two closest border points from the corresponding two clusters. Higher edges imply that the corresponding two clusters are farther apart while lower edges or unclear edges imply that clusters are not that distinct from each other.

If we observe multiple troughs in a reachability plot, as illustrated in reachability plot of Data 1 in Figure~\ref{data12}, this indicates existence of distinct clusters, i.e., the corresponding dataset has natural clusters. We call this type of structure \textit{natural cluster structure} (\textit{data structure 1}) and aim to identify these distinct clusters.
If we only observe one trough or no clear through in the reachability plot (e.g., reachability plot of Data 2 in Figure~\ref{data12}), this indicates that distinct clusters do not exist. Then we will investigate whether data points in the continuous covariate space are sufficiently heterogeneous to be further partitioned.
We use consensus clustering framework for all continuous variables to access the possible heterogeneity by checking the selected optimal number of clusters. If we obtain $\ge 2$ clusters in consensus clustering, this indicates that heterogeneity exists and we can obtain stable clusters through partitioning. We call this type of structure \textit{partitioned cluster structure} (\textit {data structure 2}).
If the optimal number of clusters is one from the consensus clustering results, this indicates that continuous part of the data is highly homogeneous and cannot be further partitioned. We call this type of structure \textit{homogeneous structure} (\textit{data structure 3}).

\subsubsection{Variable selection}
After identifying the data structure, we conduct data structure tailored variables selection.

Under the \textit{natural cluster structure}, distinct clusters can be determined by continuous variables. Therefore, we would like to select those having high contributions. As shown in Figure~\ref{fig:flow}, we apply sparse K-means on all continuous variables and keep those with high weights (suggestions of the weight threshold can be found in Section~\ref{subsec:iden}). Number of clusters under this structure can be determined by the number of troughs in the reachability plot. Next, we calculate Cramer's V between each categorical variable and the cluster membership obtained from sparse K-means. We will only select categorical variables with high Cramer's V values. Cramer's V has been used to measure the association between nominal variables. It ranges from 0 to 1. A larger number indicates a stronger association, vice versa. Unlike the \textit{p}-value, Cramer's V is not affected by the sample size. Researchers suggested the use of 0.3 as the cutoff value, namely Cramer's V larger than 0.3 indicates a moderate to strong association.

Under the \textit{partitioned cluster structure}, distinct clusters do not exist; however, covariate space of all continuous variables are sufficiently heterogeneous to be further partitioned. This structure indicates that all of the continuous variables together contribute to heterogeneity but none of them has the driving influence. Therefore, we keep all continuous variables and run consensus K-means to select the optimal number of clusters. Next, we calculate Cramer's V between each categorical variable and the cluster membership obtained from consensus K-means. We will only select categorical variables with high Cramer's V values.

Under the \textit{homogeneous structure}, no distinct cluster exists and we are not able to further partition continuous covariate space into $\ge 2$ homogeneous subgroups. Therefore, we dropped all continuous variables as they are non-distinguishable across clusters. Next, we calculate pairwise Cramer's V values among categorical variables and only select pairs with high Cramer's V values.

\subsection{Clustering step (Step 2)}
After variables with high contributions are selected, we proceed to the final clustering step. This step is the same across all data structures. We calculate the dissimilarities between subjects using our proposed dissimilarity measure, a modified version of the Gower distance. Assume that the first $h$ variables are continuous and the rest are categorical. Our proposed dissimilarity between subjects $i$ and $i^{\prime}$ is defined as:
$$d(\mathbf{x}_i,\mathbf{x}_{i^{\prime}})= \sum_{j=1}^{p}\frac{d_j(\mathbf{x}_i,\mathbf{x}_{i^{\prime}})}{\sum_{o\neq o^{\prime}}d_j(\mathbf{x}_o,\mathbf{x}_{o^{\prime}})},$$
where
\[ d_j(\mathbf{x}_i,\mathbf{x}_{i^{\prime}}) =
\begin{cases}
\frac{|x_{ij}-x_{i^{\prime}j}|}{max(\mathbf{X}_j)-min(\mathbf{X}_j)} & \quad \text{if } j \in \{1, 2, \dots, h\}\\
I(x_{ij} \neq x_{i^{\prime}j}) & \quad \text{if } j \in \{h+1, h+2, \dots, p\}
\end{cases}
\]

$$max(\mathbf{X}_j)=x_{i^\star j} \text{ if } x_{i^\star j}\geqslant x_{ij} \text{ for all } i,$$
$$min(\mathbf{X}_j)=x_{i^\star j} \text{ if } x_{i^\star j}\leqslant x_{ij} \text{ for all } i.$$

Our modification is based on the idea of standardization to avoid variables with high variability be extremely influential to clustering results. It is motivated by the definition of Gower distance for categorical variables as they receive extreme dissimilarity values $0$ or $1$, which could exhibit high variability. This allows them to exert greater influence in the clustering results even if they are less informative than the continuous ones.

Below we show how our modification on dissimilarities is analogous to the standardization on continuous variables. Standardized squared Euclidean distance between subjects $i$ and $i^{\prime}$ with respect to a continuous variable $j$ is:
$$d_j^2(\mathbf{x}_i,\mathbf{x}_{i^{\prime}})=\left\{ \frac{x_{ij}}{sd(\mathbf{X}_j)}-\frac{x_{i^{\prime}j}}{sd(\mathbf{X}_j)} \right\}^2$$
which can be re-written as:
$$d_j^2(\mathbf{x}_i,\mathbf{x}_{i^{\prime}})=\frac{(x_{ij}-x_{i^{\prime}j})^2}{n^{-2}\sum_{o\neq o^{\prime}}(x_{oj}-x_{o^{\prime}j})^2},$$
where the numerator is the original squared Euclidean distance, the denominator is proportional to the sum of all pairwise distances. We adopt this idea to standardize the Gower distance, namely we divide the original Gower distance of variable $j$ by sum of all pairwise Gower distance of variable $j$ as shown above.

If after the pre-processing step all selected variables are continuous, we can just apply usual clustering methods to obtain the final clustering results.

\subsection{Parameters selection}

In this section we provide general suggestions on the selection of (1) the optimal number of clusters; (2) continuous variables under \textit{natural cluster structure}.

\subsubsection{Number of clusters}
Under the \textit{natural cluster structure}, the number of clusters can be decided by the number of troughs in the reachability plot. Under the \textit{partitioned cluster structure}, the number of clusters can be selected from the results of the consensus clustering. Under the \textit{homogeneous structure}, we only select categorical variables in determining cluster membership. Hence we suggest constructing a dissimilarity matrix using our proposed dissimilarity measure and then plot the number of clusters against the corresponding within-cluster sum of dissimilarities. In this plot, we look for an \textit{elbow} for the optimal number of clusters.

\subsubsection{Selecting continuous variables under the natural cluster structure}
\label{subsec:iden}
Selection of the continuous variables with high weights under the \textit{natural cluster structure} could be subjective because of the choice of the weight threshold. We suggest applying sparse K-means for continuous part of each bootstrapping data set and then calculate the between-cluster sum of squares (BCSS). We then order these variables by their median BCSS from the smallest to the largest and plot the median (with $2.5^{th}$ quantile and $97.5^{th}$ quantile interval) of BCSS. Then we drop variables whose BCSS values are small or far away from the others. Our suggestion here is a heuristic one. Users can always incorporate other information and make their own judgements.

\begin{landscape}
\begingroup
\renewcommand{\arraystretch}{0.9}
\begin{table}
\begin{adjustbox}{center}
\tiny
\centering
\begin{tabular}{c c c c c c c}
\toprule
 \textbf{Variable} & \textbf{Cluster$^a$} & \textbf{Sim 1(a)} & \textbf{Sim 1(b)} & \textbf{Sim 2(a)} & \textbf{Sim 2(b)} & \textbf{Sim 3}\\
\midrule
  & 1 & $\bm{N(-2, 2)}^b$ &$\bm{N(-2, 2)}$ & $\bm{N(-2, 2)}$ & $\bm{Beta(0.1,5)}$ & \\
 $x_1$ & 2 & $\bm{N(2, 2)}$ & $\bm{N(-1, 2)}$ &$\bm{N(2, 2)}$ & $\bm{Beta(0.1,5)+0.3}$& $N(0, 0.5)$\\
 & 3 & $\bm{N(6, 2)}$ & $\bm{N(0, 2)}$&$\bm{N(6, 2)}$ &$\bm{Beta(0.1,5)+0.5}$ & \\
\midrule
  & 1 & $\bm{N(20, 1)}$ &$\bm{N(20, 1)}$ &$\bm{N(20, 1)}$ &$\bm{Beta(0.2,5)}$ & \\
 $x_2$ & 2 & $\bm{N(25, 1)}$ &$\bm{N(24, 1)}$ &$\bm{N(25, 1)}$ &$\bm{Beta(0.1,5)+0.3}$ &$N(-3, 1)$\\
 & 3 & $\bm{N(18, 1)}$ &$\bm{N(21, 1)}$ &$\bm{N(18, 1)}$ &$\bm{Beta(0.1,5)+0.5}$ & \\
\midrule
 & 1 & $\bm{N(0, 1)}$ &$\bm{N(5, 1)}$ & $\bm{N(0, 1)}$&$\bm{Beta(0.2,3)}$ & \\
 $x_3$ & 2 & $\bm{N(-7, 1)}$ &$\bm{N(8, 1)}$ &$\bm{N(-7, 1)}$ &$\bm{Beta(0.2,3)+0.3}$ &$N(4, 2)$\\
 & 3 & $\bm{N(4, 1)}$ &$\bm{N(7, 1)}$ &$\bm{N(4, 1)}$ &$\bm{Beta(0.2,3)+0.5}$ & \\
 \midrule
  & 1 & & & &$\bm{Beta(0.1,3)}$ & \\
 $x_4$ & 2 & $N(0, 1)$ & $N(0, 1)$&$N(0, 1)$ & $\bm{Beta(0.1,3)+0.3}$&$N(0, 1)$\\
 & 3 & & & &$\bm{Beta(0.2,3)+0.5}$ & \\
 \midrule
 & 1 & $\bm{M(0.1, 0.1, 0.8)}$ & &$M(0.3, 0.3, 0.4)$ & & $\bm{M(0.05, 0.05, 0.9)}$ \\
 $x_5$ & 2 & $\bm{M(0.1, 0.8, 0.1)}$ &$N(40, 1)$ & $M(0.3, 0.3, 0.4)$& $N(0, 0.01)$& $\bm{M(0.05, 0.9, 0.05)}$\\
 & 3 & $\bm{M(0.8, 0.1, 0.1)}$ & &$M(0.4, 0.3, 0.3)$ & & $\bm{M(0.9, 0.05, 0.05)}$\\
 \midrule
 & 1 & &$\bm{N(-1, 1)}$ & &$M(0.3, 0.3, 0.4)$ & $M(0.3, 0.3, 0.4)$\\
 $x_6$ & 2 & & $\bm{N(1, 1)}$ & &$M(0.4, 0.3, 0.3)$  &$M(0.4, 0.3, 0.3)$\\
 & 3 & &$\bm{N(-2, 1)}$ & & $M(0.3, 0.4, 0.3)$ & $M(0.3, 0.4, 0.3)$\\
 \midrule
 & 1 & &$\bm{N(0, 1)}$ & & & $\bm{M(0.9, 0.05, 0.05)}$\\
 $x_7$ & 2 & &$\bm{N(-1, 1)}$ & &$M(0.3, 0.3, 0.4)$ &$\bm{M(0.05, 0.9, 0.05)}$\\
 & 3 & &$\bm{N(2, 1)}$ & & & $\bm{M(0.05, 0.05, 0.9)}$\\
 \midrule
 & 1 & &$\bm{N(2, 1)}$ & & & \\
 $x_8$ & 2 & &$\bm{N(1, 1)}$ & &$M(0.3, 0.3, 0.4)$  &\\
 & 3 & &$\bm{N(0, 1)}$ & & & \\
 \midrule
 $x_9 \sim x_{11}$ & 1,2,3 & &$N(0, 1)$ & & &\\
 \midrule
 & 1 & &$M(0.3, 0.3, 0.4)$ & & & \\
 $x_{12}$ & 2 & &$M(0.4, 0.3, 0.3)$ & & &\\
 & 3 & &$M(0.3, 0.4, 0.3)$ & & & \\
 \midrule
 & 1 & &$\bm{M(0.9, 0.05, 0.05)}$ & & & \\
 $x_{13}$ & 2 & &$\bm{M(0.05, 0.9, 0.05)}$ & & &\\
 & 3 & &$\bm{M(0.05, 0.05, 0.9)}$ & & & \\
 \midrule
 & 1 & &$\bm{M(0.05, 0.05, 0.9)}$ & & & \\
 $x_{14}$ & 2 & &$\bm{M(0.05, 0.9, 0.05)}$ & & &\\
 & 3 & &$\bm{M(0.9, 0.05, 0.05)}$ & & & \\ \midrule
  \multicolumn{7}{l}{$^a$Sample sizes for 3 clusters are 40, 40 and 120; $^b$variables with bolded distributions are important in clustering} \\ \bottomrule
 \end{tabular}
\caption{Simulation settings}
\label{tab:setting}
\end{adjustbox}
\end{table}
\endgroup
\end{landscape}

\section{Simulation studies}
\label{sec:sim}
In this section we use simulations to evaluate the performance of the HyDaP algorithm relative to the existing approaches. Assuming that there are 3 underlying true clusters with cluster sizes of 40, 40, and 120. In terms of variable importance, we considered scenarios (1) both variable types contribute to clustering, (2) only continuous variables contribute to clustering, and (3) only categorical variables contribute to clustering. In terms of data structures, all 3 data structures were covered in simulations. Details of the distributions and parameters used in these simulation settings are shown in Table ~\ref{tab:setting}.

For each setting, 500 datasets were generated. Cluster analysis was performed on each dataset using the proposed HyDaP algorithm. We compared its performance with PAM with Gower distance, K-prototypes, FMM, and PAM with FAMD distance. Since we know the true cluster labels, the adjusted rand index (ARI) was calculated and used to evaluate the performances of different methods. ARI is used to measure the agreement between two nominal variables. Its largest value is 1 indicating perfect agreement and its smallest value is close to 0 indicating no agreement. For the purpose of evaluating clustering performance in simulations, higher ARI values indicate better agreement with true cluster labels and hence better performance. The reachability plot for each setting is illustrated in Figure~\ref{fig:figure2}. Table~\ref{tab:step1} summarizes the results of the pre-processing step of the HyDaP algorithm. The clustering performance with respect to ARI across all simulation settings is shown in Table~\ref{tab:perform}.
To examine the impact of conditional correlation on clustering performance, each simulation setting was imbued with a pairwise correlation of 0.4 conditional on true cluster labels. Results are shown in Table~\ref{tab:corr}. Median along with the $2.5^{th}$ and $97.5^{th}$ percentiles were reported for all statistics.

\subsection{Setting 1: Both types of variables contribute to clustering}
\subsubsection{Natural cluster structure}
In simulation 1(a), we simulated a total of 5 variables: 4 continuous and 1 categorical. All except one continuous variable truly contribute to clustering. The sole categorical variable also contributes to clustering.

\begin{figure}
  \begin{subfigure}[b]{0.5\linewidth}
    \centering
    \includegraphics[width=0.75\linewidth]{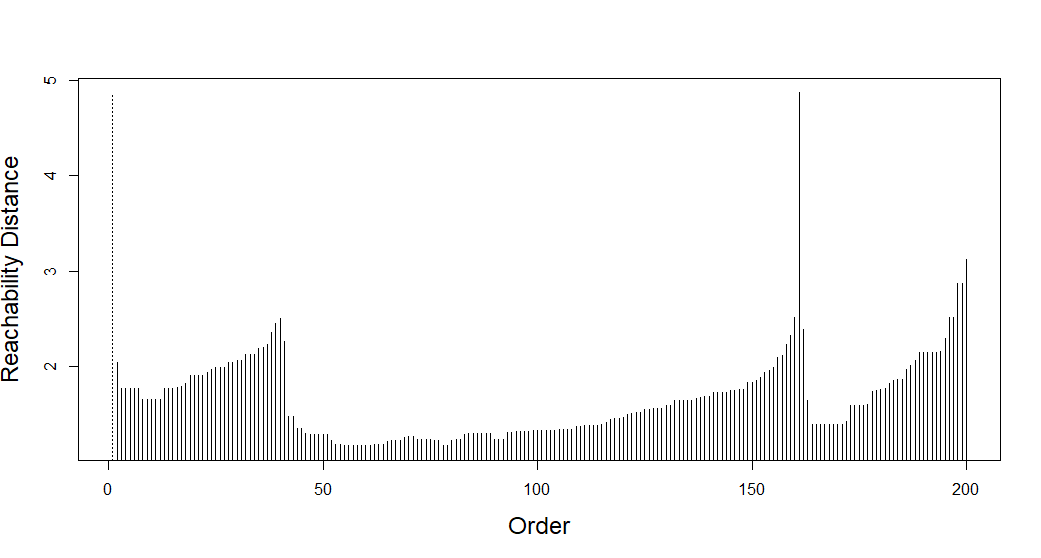}
    \caption{Sim 1(a)}
    \label{fig:sub21}
    \vspace{4ex}
  \end{subfigure}
  \begin{subfigure}[b]{0.5\linewidth}
    \centering
    \includegraphics[width=0.75\linewidth]{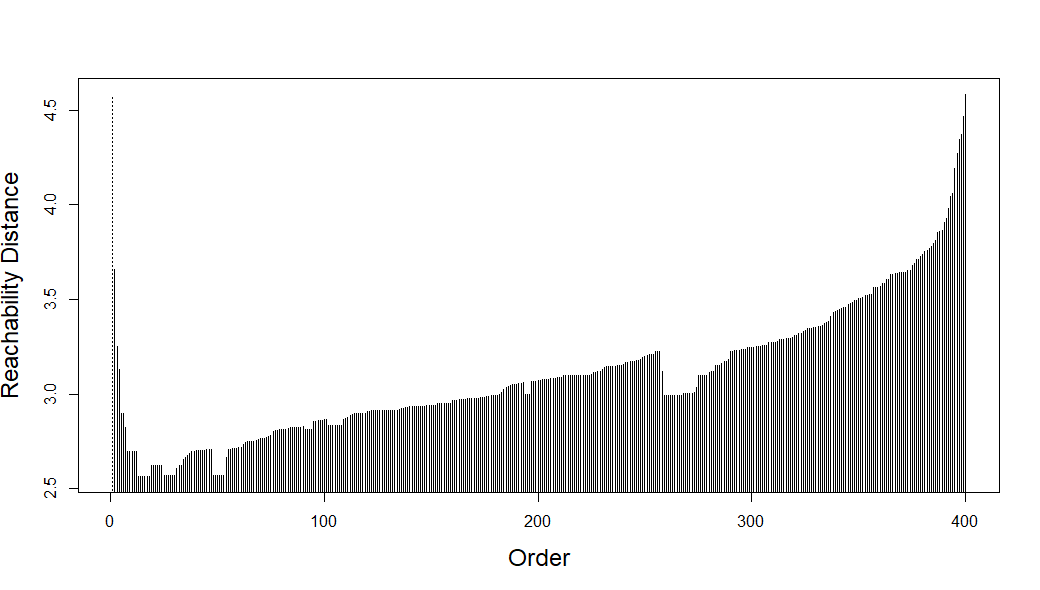}
    \caption{Sim 1(b)}
    \label{fig:sub23}
    \vspace{4ex}
  \end{subfigure}
  \begin{subfigure}[b]{0.5\linewidth}
    \centering
    \includegraphics[width=0.75\linewidth]{figure6}
    \caption{Sim 2(a)}
    \label{fig:sub25}
  \end{subfigure}
  \begin{subfigure}[b]{0.5\linewidth}
    \centering
    \includegraphics[width=0.75\linewidth]{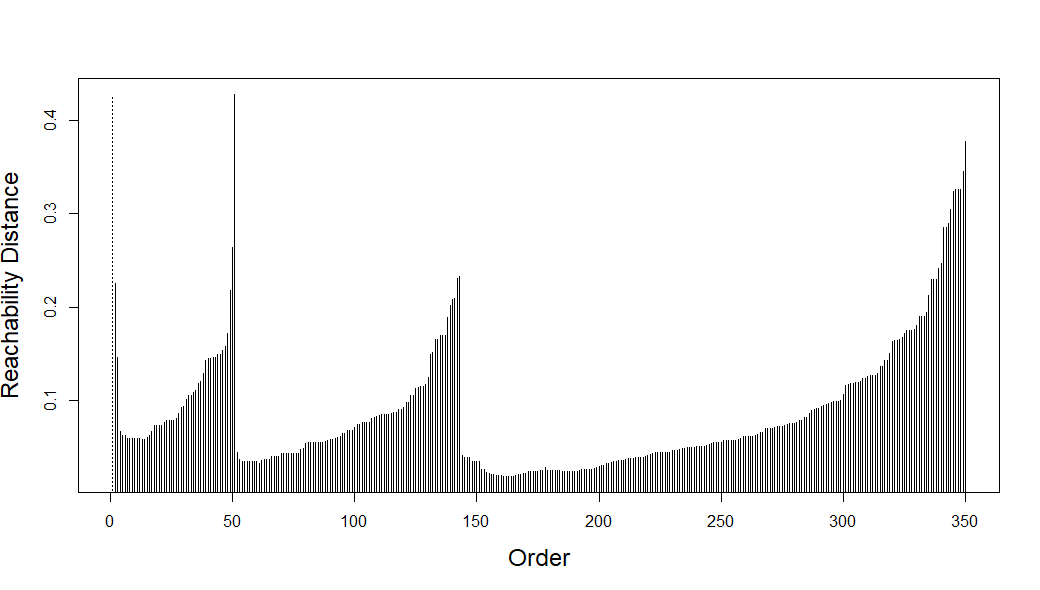}
    \caption{Sim 2(b)}
    \label{fig:sub22}
  \end{subfigure}
  \begin{subfigure}[b]{0.5\linewidth}
    \centering
    \includegraphics[width=0.75\linewidth]{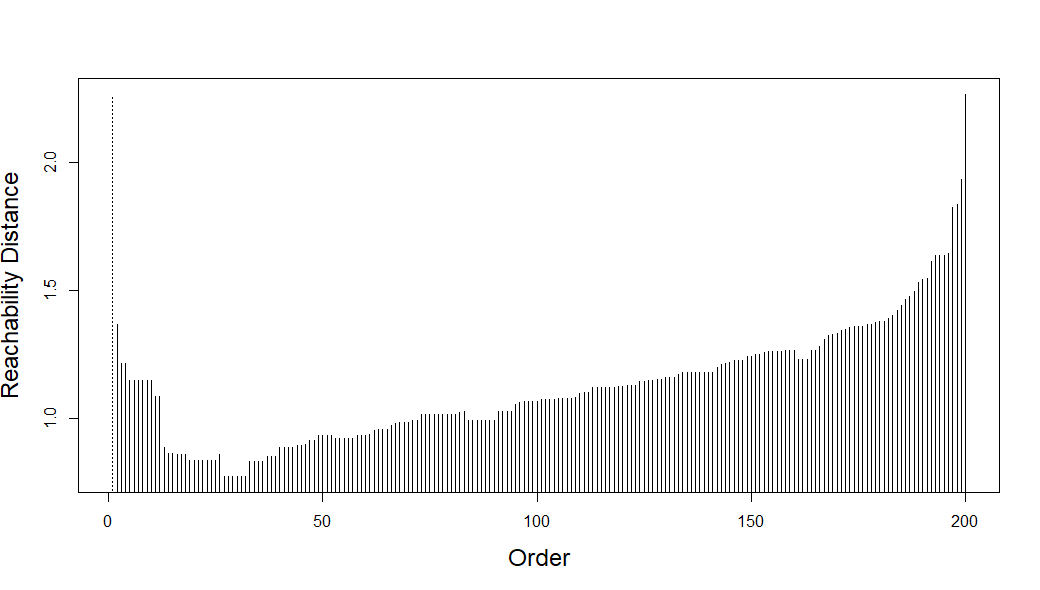}
    \caption{Sim 3}
    \label{fig:sub24}
  \end{subfigure}
  \caption{Reachability plots in different simulation settings}
  \label{fig:figure2}
\end{figure}

In Step 1 of the HyDaP algorithm, the reachability plot (Figure~\ref{fig:sub21}) indicated 3 clusters. Therefore, this setting has natural cluster structure as 3 distinct clusters exist. Table~\ref{tab:step1} shows the very low contribution of $x_4$ from the sparse K-means and the strong association between $x_5$ and the clusters identified by the sparse K-means. We dropped $x_4$ and kept all the others.

In Step 2, we applied PAM along with the proposed dissimilarity measure on the selected variables from Step 1: $x_1$, $x_2$, $x_3$, and $x_5$.

As shown in Table~\ref{tab:perform}, HyDaP algorithm performed very well [ARI: 0.97 (0.92, 1.00)]. Although K-prototypes (ARI: 1.00 [0.96, 1.00]) and FMM (ARI: 1.00 [0.98, 1.00]) both performed slightly better, our HyDaP algorithm was able to identify important variables. PAM with Gower distance (ARI: 0.70 [0.58, 0.80]) and PAM with FAMD distance (ARI: 0.78 [0.66, 0.89]) performed poorly. This is not surprising of the results using the Gower distance since it tends to downplay contributions of continuous variables, although in this setting continuous variables $x_1$ to $x_3$ all have large contributions to clustering.

\subsubsection{Partitioned cluster structure}
In simulation 1(b), we simulated a total of 14 variables: 11 continuous and 3 categorical. Six out of eleven continuous variables truly contribute to clustering; two out of three categorical variables contribute to clustering.

In Step 1 of the HyDaP algorithm, Figure~\ref{fig:sub23} indicated that no natural clusters exists. After conducting consensus K-means, we chose 3 as the optimal number of clusters as its corresponding cluster-consensus values were the largest. Thus, a partitioned cluster structure was identified. All continuous variables were retained for the next step. Variable $x_{12}$ was dropped because of its small Cramer's V with cluster assignments obtained in consensus K-means.

In Step 2, PAM with proposed dissimilarity measure was applied on $x_1$, $x_2$,\dots, $x_{11}$, $x_{13}$, and $x_{14}$ to obtain final results.

Performance of the HyDaP algorithm is satisfactory (ARI 0.95 [0.87, 1.00]). Although it was unable to eliminate continuous variables that are purely noise, the HyDaP algorithm revealed that no continuous variable has driving effect but all of them together lead to heterogeneity in the feature space spanned by all of these continuous variables. In this setting, K-prototypes (ARI: 0.93 [0.79, 1.00]) and PAM with FAMD distance (ARI: 0.93 [0.84, 0.98]) also worked well while performance of FMM varied widely from sample to sample (ARI: 0.98 [0.44, 1.00]). PAM with Gower distance did not perform as well as others (ARI: 0.87 [0.76, 0.96]). This is because a noise categorical variable $x_{12}$ was included and Gower distance tends to amplify its contribution.

\begin{table}[]
\begin{adjustbox}{center}
\footnotesize
\centering
\resizebox{0.95\textwidth}{!}{%
\begin{tabular}{@{}lcllc@{}}
\toprule
\multicolumn{1}{c}{Sim 1(a)} & Sim 1(b) & \multicolumn{1}{c}{Sim 2(a)} & \multicolumn{1}{c}{Sim 2(b)} & Sim 3 \\ \midrule
\multicolumn{5}{c}{Data Structure} \\
\multicolumn{1}{c}{1} & 2 & \multicolumn{1}{c}{1} & \multicolumn{1}{c}{1} & 3 \\ \midrule
\multicolumn{1}{c}{Weight} & - & \multicolumn{1}{c}{Weight} & \multicolumn{1}{c}{Weight} & - \\
$x_1$: 0.49 (0.46, 0.52) & \multirow{5}{*}{\begin{tabular}[c]{@{}c@{}}Keep all\\ continuous\\ variables\end{tabular}} & $x_1$: 0.49 (0.46, 0.52) & $x_1$: 0.54 (0.51, 0.57) & \multirow{5}{*}{\begin{tabular}[c]{@{}c@{}}Drop all\\ continuous\\ variables\end{tabular}} \\
$x_2$: 0.59 (0.58, 0.61) &  & $x_2$: 0.59 (0.58, 0.61) & $x_2$: 0.51 (0.48, 0.54) &  \\
$x_3$: 0.64 (0.62, 0.65) &  & $x_3$: 0.64 (0.62, 0.65) & $x_3$: 0.44 (0.38, 0.48) &  \\
$x_4$: 0.00 (0.00, 0.02) &  & $x_4$: 0.00 (0.00, 0.02) & $x_4$: 0.50 (0.45, 0.54) &  \\
 &  &  & $x_5$: 0.00 (0.00, 0.02) &  \\ \midrule
\multicolumn{1}{c}{Cramer's V} & Cramer's V & \multicolumn{1}{c}{Cramer's V} & \multicolumn{1}{c}{Cramer's V} & \begin{tabular}[c]{@{}c@{}}Pairwise\\ Cramer's V\end{tabular} \\
$x_5$: 0.66 (0.57, 0.75) & \multicolumn{1}{l}{$x_{12}$: 0.12 (0.05, 0.21)} & $x_5$: 0.13 (0.06, 0.21) & $x_6$: 0.12 (0.05, 0.21) & \multicolumn{1}{l}{$x_5$ $x_6$: 0.12 (0.04, 0.20)} \\
 & \multicolumn{1}{l}{$x_{13}$: 0.77 (0.69, 0.85)} &  & $x_7$: 0.09 (0.04, 0.17) & \multicolumn{1}{l}{$x_5$ $x_7$: 0.69 (0.60, 0.77)} \\
 & \multicolumn{1}{l}{$x_{14}$: 0.78 (0.69, 0.86)} &  & $x_8$: 0.09 (0.04, 0.17) & \multicolumn{1}{l}{$x_6$ $x_7$: 0.12 (0.04, 0.19)} \\ \midrule
 \multicolumn{5}{l}{$^a$all weights and cramer's v values are presented in the form of median (2.5th percentile, 97.5th percentile).} \\ \bottomrule
\end{tabular}%
}
\caption{Results from pre-processing step in different simulation settings }
\label{tab:step1}
\end{adjustbox}
\end{table}

\subsection{Setting 2: Only continuous variables contribute to clustering}
\subsubsection{Natural cluster structure}

In simulation 2(a), we simulated a total of 5 variables: 4 continuous and 1 categorical. This setting is the same as simulation 1(a) except that the sole categorical variable does not contribute to clustering.

In Step 1 of the HyDaP algorithm, $x_4$ was dropped due to its low contribution in the sparse K-means. Table~\ref{tab:step1} shows a weak association between the categorical variable $x_5$ and clusters identified by the sparse K-means.

In Step 2, we applied the sparse K-means on $x_1$, $x_2$, and $x_3$ as they are all continuous variables.

In this setting, the HyDaP algorithm (ARI: 0.98 [0.94, 1.00]) and K-prototypes (ARI: 0.98 [0.92, 1.00]) both worked well. There were a few simulation runs the performance of FMM was not satisfactory (ARI: 1.00 [0.56, 1.00]). PAM with Gower distance (ARI: 0.01 [-0.01, 0.04]) and PAM with FAMD distance (ARI: 0.09 [-0.01, 0.44]) performed extremely poor. As mentioned in simulation 1(b), Gower distance tends to amplify the contributions of the categorical variables. Meanwhile, FAMD was not originally designed for clustering.

\subsubsection{Natural cluster structure}

In simulation 2(b), we simulated a total of 8 variables: 5 continuous and 3 categorical. Four out of five continuous variables truly contribute to clustering and follow highly skewed distributions. None of the categorical variables contributes to clustering.

In Step 1 of the HyDaP algorithm, Figure~\ref{fig:sub22} shows 3 distinct clusters and hence this setting was identified as natural cluster structure. We dropped $x_5$ because of its small contribution to clustering as shown in Table~\ref{tab:step1}. All categorical variables were dropped as well given their weak associations with clusters obtained in the sparse K-means.

In Step 2, we applied the sparse K-means on $x_1$, $x_2$, and $x_3$ since they are all continuous variables.

In this setting, the HyDaP algorithm performed the best (ARI: 0.98 [0.92, 1.00]). PAM with Gower distance (ARI: 0.23 [0.00, 0.34]), K-prototypes (ARI: 0.58 [0.38, 0.99]), FMM (ARI: 0.41 [0.33, 0.58]), and PAM with FAMD distance (ARI: 0.34 [0.08, 0.39]) all performed poorly. This was expected for FMM because most of the continuous variables were not normally distributed conditional on the true cluster labels.

\subsection{Setting 3: only categorical variables contribute to clustering}
\subsubsection{Homogeneous structure}

In simulation 3, we simulated a total of 7 variables: 4 continuous and 3 categorical. None of the continuous variables truly contributes to clustering. Two out of three categorical variables contribute to clustering.

In Step 1 of the HyDaP algorithm, Figure~\ref{fig:sub24} indicates no natural clusters exist. After conducting consensus K-means, the optimal number of clusters chosen was 1 because cluster-consensus values were low for all numbers of clusters. Hence this was identified as homogeneous structure. All continuous variables were dropped but categorical variables $x_5$ and $x_7$ were kept due to their strong association with each other as shown in Table~\ref{tab:step1}.

In Step 2, PAM with proposed dissimilarity measure was applied on $x_5$ and $x_7$.

In this setting, the HyDaP algorithm performed the best (ARI: 0.75 [0.63, 0.85]) and K-prototypes did the worst (ARI: 0.17 [-0.01, 0.26]). Performance of PAM with Gower distance (ARI: 0.71 [0.31, 0.84]), FMM [ARI: 0.72 (0.56, 0.85)] and PAM with FAMD distance (ARI: 0.73 [0.22, 0.84]) were similar.

\begin{table}
\footnotesize
\centering
\begin{adjustbox}{center}
\resizebox{0.95\textwidth}{!}{%
\begin{tabular}{@{}cccccc@{}}
\toprule
\textbf{Clustering Method} & \multicolumn{5}{c}{\textbf{ARI, median (2.5th percentile, 97.5th percentile)}} \\ \midrule
 & Sim 1(a) & Sim 1(b) & Sim 2(a) & Sim 2(b) & Sim 3 \\ \cmidrule(l){2-6}
\footnotesize{HyDaP} & \begin{tabular}[c]{@{}c@{}}0.97 \\ (0.92, 1.00)\end{tabular} & \begin{tabular}[c]{@{}c@{}}0.95 \\ (0.87, 1.00)\end{tabular} & \begin{tabular}[c]{@{}c@{}}0.98\\ (0.94, 1.00)\end{tabular} & \begin{tabular}[c]{@{}c@{}}0.98 \\ (0.92, 1.00)\end{tabular} & \begin{tabular}[c]{@{}c@{}}0.75 \\ (0.63, 0.85)\end{tabular} \\
\footnotesize{PAM + Gower distance} & \begin{tabular}[c]{@{}c@{}}0.70 \\ (0.58, 0.80)\end{tabular} & \begin{tabular}[c]{@{}c@{}}0.87 \\ (0.76, 0.96)\end{tabular} & \begin{tabular}[c]{@{}c@{}}0.01 \\ (-0.01, 0.04)\end{tabular} & \begin{tabular}[c]{@{}c@{}}0.23 \\ (0.00, 0.34)\end{tabular} & \begin{tabular}[c]{@{}c@{}}0.71 \\ (0.31, 0.84)\end{tabular} \\
\footnotesize{K-prototypes} & \begin{tabular}[c]{@{}c@{}}1.00 \\ (0.96, 1.00)\end{tabular} & \begin{tabular}[c]{@{}c@{}}0.93\\ (0.79, 0.95)\end{tabular} & \begin{tabular}[c]{@{}c@{}}0.98 \\ (0.92, 1.00)\end{tabular} & \begin{tabular}[c]{@{}c@{}}0.58 \\ (0.38, 0.99)\end{tabular} & \begin{tabular}[c]{@{}c@{}}0.17 \\ (-0.01, 0.26)\end{tabular} \\
\footnotesize{Finite mixture model} & \begin{tabular}[c]{@{}c@{}}1.00 \\ (0.98, 1.00)\end{tabular} & \begin{tabular}[c]{@{}c@{}}0.98\\ (0.44, 1.00)\end{tabular} & \begin{tabular}[c]{@{}c@{}}1.00 \\ (0.56, 1.00)\end{tabular} & \begin{tabular}[c]{@{}c@{}}0.41 \\ (0.33, 0.58)\end{tabular} & \begin{tabular}[c]{@{}c@{}}0.72 \\ (0.56, 0.85)\end{tabular} \\
\footnotesize{PAM + FAMD distance} & \begin{tabular}[c]{@{}c@{}}0.78 \\ (0.66, 0.89)\end{tabular} & \begin{tabular}[c]{@{}c@{}}0.93 \\ (0.84, 0.98)\end{tabular} & \begin{tabular}[c]{@{}c@{}}0.09 \\ (-0.01, 0.44)\end{tabular} & \begin{tabular}[c]{@{}c@{}}0.34 \\ (0.08, 0.39)\end{tabular} & \begin{tabular}[c]{@{}c@{}}0.73 \\ (0.22, 0.84)\end{tabular} \\
\bottomrule
\end{tabular}%
}
\caption{Performance comparison under different simulation settings}
\label{tab:perform}
\end{adjustbox}
\end{table}

\subsection{Variables are conditionally correlated}

To assess the impact of within-cluster correlation, simulations for each of the 5 settings above was repeated with pairwise correlation of 0.4 for all continuous variables conditional on true cluster labels. As summarized in Table~\ref{tab:corr}, within-cluster correlation had little to no impact on the performance of the HyDaP algorithm, PAM with Gower distance, K-prototypes, and PAM with FAMD distance.
In some situations, it led to worse performance of FMM. This is expected since FMM assumes conditional independency, namely all variables are independent with each other conditional on clusters labels. However, we did observe that in simulation 3 when none of the continuous variables contributes to clustering, the optimal number of clusters selected by the consensus K-means was 2 instead of 3 (figures not shown here).
This is understandable since all pairs of continuous variables are correlated given true cluster labels, therefore, they share a lot of common information. To some extent we can use only one of them without losing much information as all others as redundant. For any single continuous variable we can potentially divide it into 2 subgroups that have some differences. But this does not essentially mean these 2 subgroups can be viewed as 2 clusters.
Therefore, if we observe that 2 is the optimal number of clusters in consensus clustering results and most pairs of continuous variables have high conditional correlations, we should be cautious. One suggestion is that we could try to look for continuous variables that have similar clinical meanings e.g., Aspartate Aminotransferase (AST) and Alanine Aminotransferase (ALT), since these variables are very likely to have high correlations within clusters. For these variables we can only keep one of them in clustering to avoid such situation.

\begin{table}[]
\footnotesize
\centering
\begin{adjustbox}{center}
\resizebox{0.95\textwidth}{!}{%
\begin{tabular}{@{}cccccc@{}}
\toprule
\textbf{Clustering Method} & \multicolumn{5}{c}{\textbf{ARI, median (2.5th percentile, 97.5th percentile)}} \\ \midrule
 & Sim 1(a) & Sim 1(b) & Sim 2(a) & Sim 2(b) & Sim 3 \\ \cmidrule(l){2-6}
\footnotesize{HyDaP} & \begin{tabular}[c]{@{}c@{}}0.97 \\ (0.92, 1.00)\end{tabular} & \begin{tabular}[c]{@{}c@{}}0.94 \\ (0.33, 1.00)\end{tabular} & \begin{tabular}[c]{@{}c@{}}1.00\\ (0.95, 1.00)\end{tabular} & \begin{tabular}[c]{@{}c@{}}1.00 \\ (0.97, 1.00)\end{tabular} & \begin{tabular}[c]{@{}c@{}}0.74 \\ (0.61, 0.83)\end{tabular} \\
\footnotesize{PAM + Gower distance} & \begin{tabular}[c]{@{}c@{}}0.71 \\ (0.59, 0.80)\end{tabular} & \begin{tabular}[c]{@{}c@{}}0.87 \\ (0.77, 0.95)\end{tabular} & \begin{tabular}[c]{@{}c@{}}0.01 \\ (-0.01, 0.04)\end{tabular} & \begin{tabular}[c]{@{}c@{}}0.33 \\ (0.21, 0.38)\end{tabular} & \begin{tabular}[c]{@{}c@{}}0.73 \\ (0.28, 0.83)\end{tabular} \\
\footnotesize{K-prototypes} & \begin{tabular}[c]{@{}c@{}}0.98 \\ (0.95, 1.00)\end{tabular} & \begin{tabular}[c]{@{}c@{}}0.62\\ (0.32, 0.98)\end{tabular} & \begin{tabular}[c]{@{}c@{}}0.97 \\ (0.89, 1.00)\end{tabular} & \begin{tabular}[c]{@{}c@{}}0.99 \\ (0.95, 1.00)\end{tabular} & \begin{tabular}[c]{@{}c@{}}0.13 \\ (-0.01, 0.23)\end{tabular} \\
\footnotesize{Finite mixture model} & \begin{tabular}[c]{@{}c@{}}1.00 \\ (0.46, 1.00)\end{tabular} & \begin{tabular}[c]{@{}c@{}}0.47\\ (0.36, 1.00)\end{tabular} & \begin{tabular}[c]{@{}c@{}}1.00 \\ (0.97, 1.00)\end{tabular} & \begin{tabular}[c]{@{}c@{}}0.58 \\ (0.46, 1.00)\end{tabular} & \begin{tabular}[c]{@{}c@{}}0.29 \\ (0.14, 0.82)\end{tabular} \\
\footnotesize{PAM + FAMD distance} & \begin{tabular}[c]{@{}c@{}}0.79 \\ (0.67, 0.88)\end{tabular} & \begin{tabular}[c]{@{}c@{}}0.93 \\ (0.34, 0.98)\end{tabular} & \begin{tabular}[c]{@{}c@{}}0.34 \\ (-0.01, 0.44)\end{tabular} & \begin{tabular}[c]{@{}c@{}}0.35 \\ (0.22, 0.41)\end{tabular} & \begin{tabular}[c]{@{}c@{}}0.72 \\ (0.22, 0.84)\end{tabular} \\ \bottomrule
\end{tabular}%
}
\caption{Performance comparison with existence of conditionally correlated variables}
\label{tab:corr}
\end{adjustbox}
\end{table}

\subsection{Simulation summary}

From the simulation studies, we found that our proposed HyDaP algorithm was consistently the top or one of the top performers across all simulation settings. Moreover, we found that (1) when categorical variables do not contribute much to clustering, PAM with Gower distance performed poorly; (2) when continuous variables follow arbitrary distributions, FMM may not perform well due to assumption violation; (3) when none of continuous variables contributes to clustering, K-prototypes may fail; (4) performance of PAM with FAMD distance was not stable across different scenarios as its distance measure is not specifically designed for clustering.

\section{Real data application}
\label{sec:real}

We used the EHR data collected from the Sepsis ENdotyping in Emergency CAre (SENECA) project to demonstrate the use of our proposed HyDaP algorithm for identifying phenotypes in patients with sepsis. The SENECA data contains 20,189 sepsis encounters collected from 12 University of Pittsburgh Medical Center (UPMC) healthcare systems from year 2010 to 2012. The goal is to explore whether clinical sepsis phenotypes are identifiable for a patient that presents at the emergency department and whether the phenotypes are associated with various clinical endpoints.
The objectives of the analysis are to select the most important variables among 30 demographic and clinical variables, and to identify homogeneous clusters (phenotypes). Twenty eight variables were continuous and 2 were categorical. Although we do not have much information about the optimal number of clusters for the data set, our clinician colleagues suggested that larger numbers of clusters are preferred.

\begin{figure}
\centering
\includegraphics[width=0.9\textwidth]{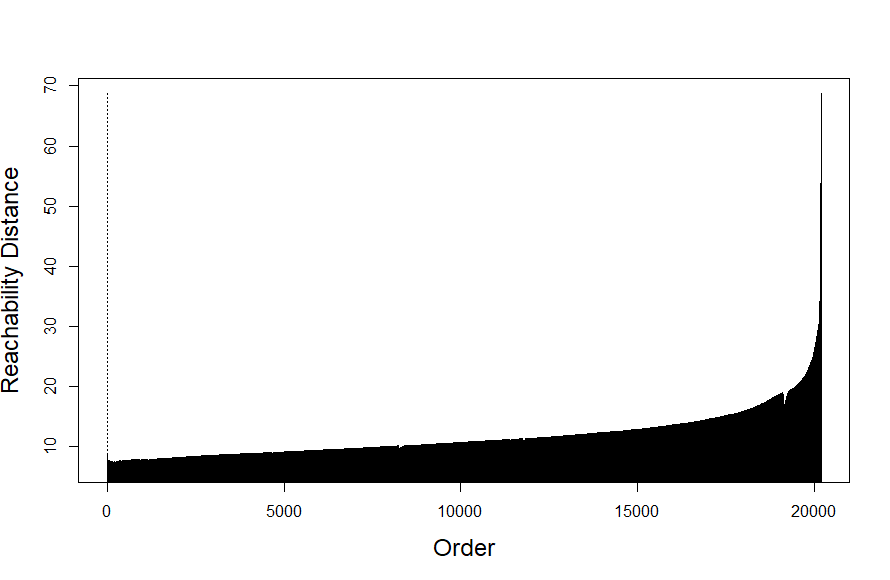}
\caption{OPTICS reachability plot for the SENECA data}
\label{seneca}
\end{figure}

\begin{figure}
\centering
\includegraphics[width=\textwidth]{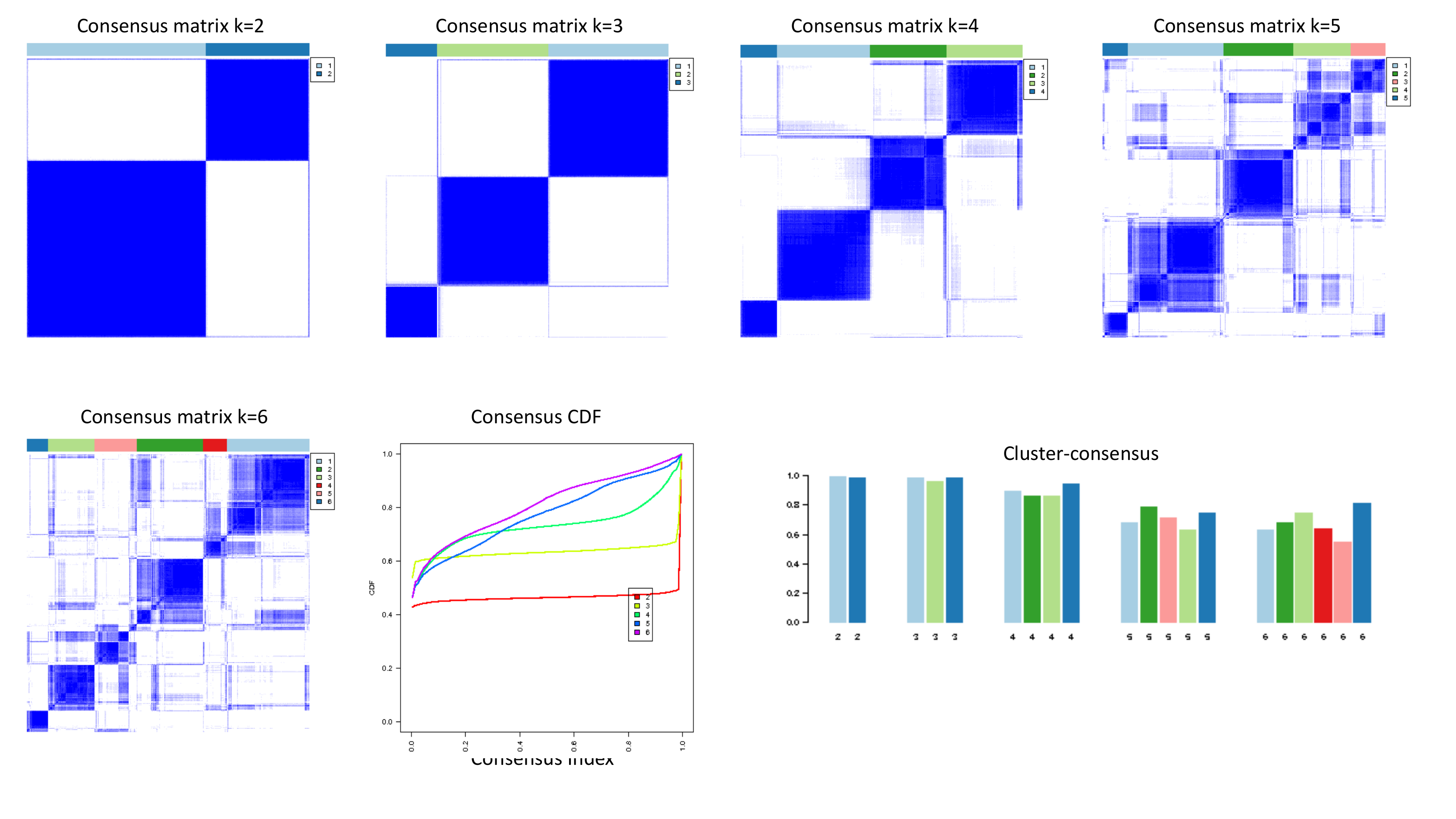}
\caption{Consensus K-means optimal number of clusters selection}
\label{consensusk}
\end{figure}

\textit{Data structure identification}: The reachability plot in Figure~\ref{seneca} indicates that there is no natural clusters in the SENECA data. Unlike the genetic data, we rarely observe natural clusters in data collected from clinical settings. We then performed the consensus K-means for all continuous variables, and the results are depicted in Figure~\ref{consensusk} suggesting that the optimal number of clusters is $4$, which indicates that the data structure of the SENECA data belongs to \textit{partitioned cluster structure}.

\textit{Variable selection}: Under \textit{partitioned cluster structure} we kept all continuous variables. For categorical variables, Cramer's V is 0.05 between gender and cluster membership from consensus clustering and it is also 0.05 between race and cluster membership. Therefore, we dropped gender and race before proceeding to the final clustering step.

\begin{figure}
\centering
\includegraphics[width=\textwidth]{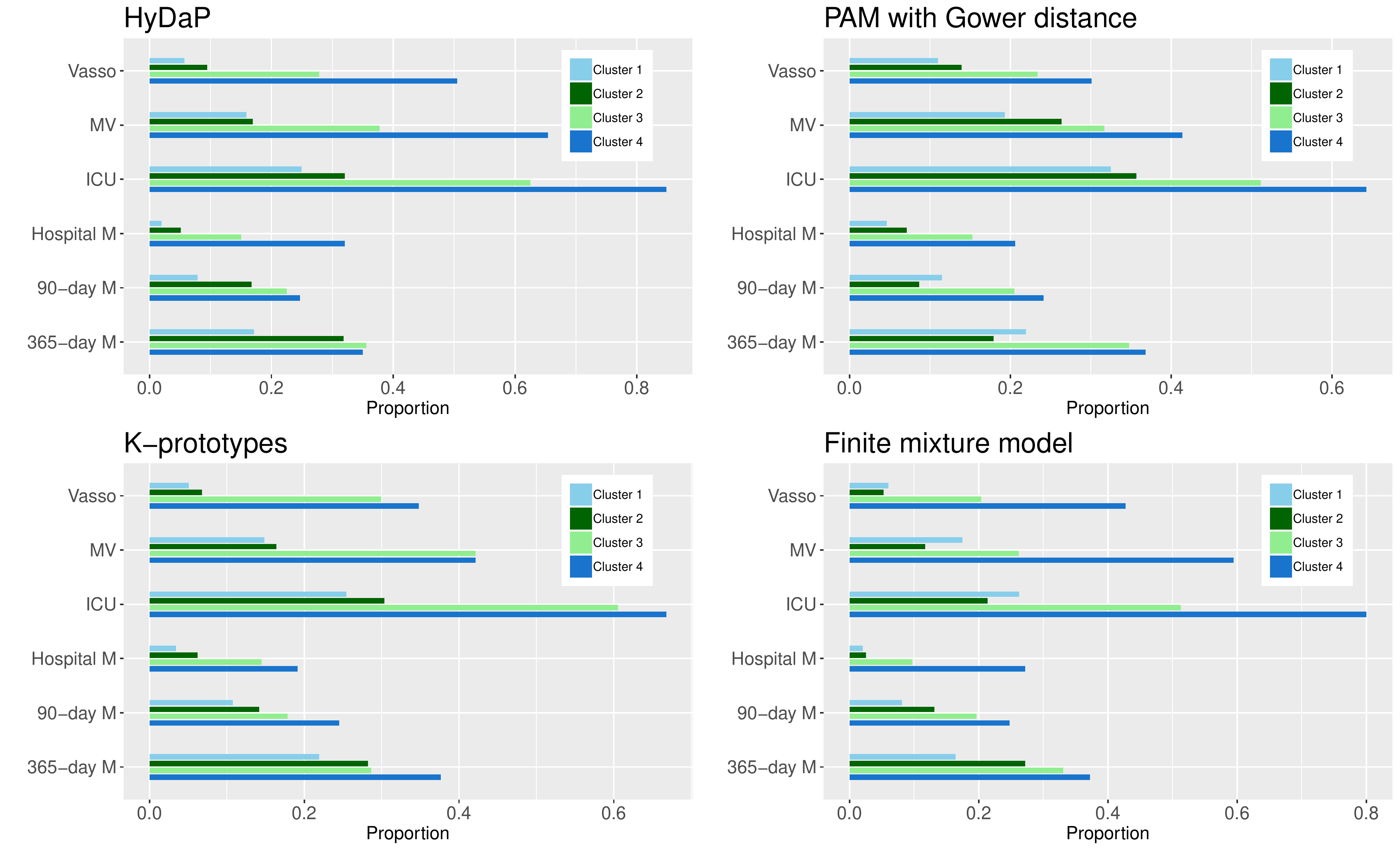}
\caption{Clinical endpoints across 4 clusters identified by different methods}
\label{fig:senecak4}
\end{figure}

\textit{Clustering step}: All categorical variables were excluded after the pre-processing step, so we took the results from the consensus K-means as our final clustering results. In terms of variable importance, all continuous variables together had contributions to the obtained partitions but none of them showed dominant impact. Neither gender nor race were important clustering variables.
We obtained 4 clusters with relatively balanced sample sizes: $6,625$, $5,512$, $5,385$, and $2,667$. Within each cluster, distributions of some important clinical endpoints are shown in top left plot of Figure~\ref{fig:senecak4}.
We can observe that Cluster 1 has the lowest proportion for all clinical endpoints while Cluster 2 has the second lowest ones. Cluster 4 has the highest proportions. With our clinician colleagues, we examined patient characteristics of the resulting clusters. We observed that sepsis patients in Cluster 1 had fewer other health issues; patients in Cluster 2 were those who were older, had multi morbidities, and renal dysfunctions; patients in Cluster 3 were those who had more inflammations and pulmonary dysfunctions; and patients in Cluster 4 were whose who had more acidosis, liver, and cardiovascular dysfunctions.

For comparison, we applied PAM with Gower distance, K-prototypes, and FMM to obtain cluster memberships assuming 4 clusters. The results are summarized in Figure~\ref{fig:senecak4}.
For PAM with Gower distance, we took a random sample of the whole SENECA data with size $5,000$ because the computation time of this algorithm was very long. After further exploration we found that gender dominated the clustering result as the proportion of male is $0.0\%$ in Cluster 1, $2.7\%$ in Cluster 4, $99.4\%$ in Cluster 2, and $99.8\%$ in Cluster 3. Note that gender was not relevant based on our proposed HyDaP algorithm.
For the K-prototypes, we found that the 4 clusters obtained were not that distinct from each other in terms of the distribution of clinical endpoints.
The 4 clusters obtained from the FMM appeared to be distinct from each other and similar to what we observed in HyDaP algorithm. However, Cluster 1 has larger proportion of patients admitted to ICU, use of mechanical ventilation and vasopressor compared to Cluster 2, but it has lower mortality rate.

\begin{figure}
\centering
\includegraphics[width=\textwidth]{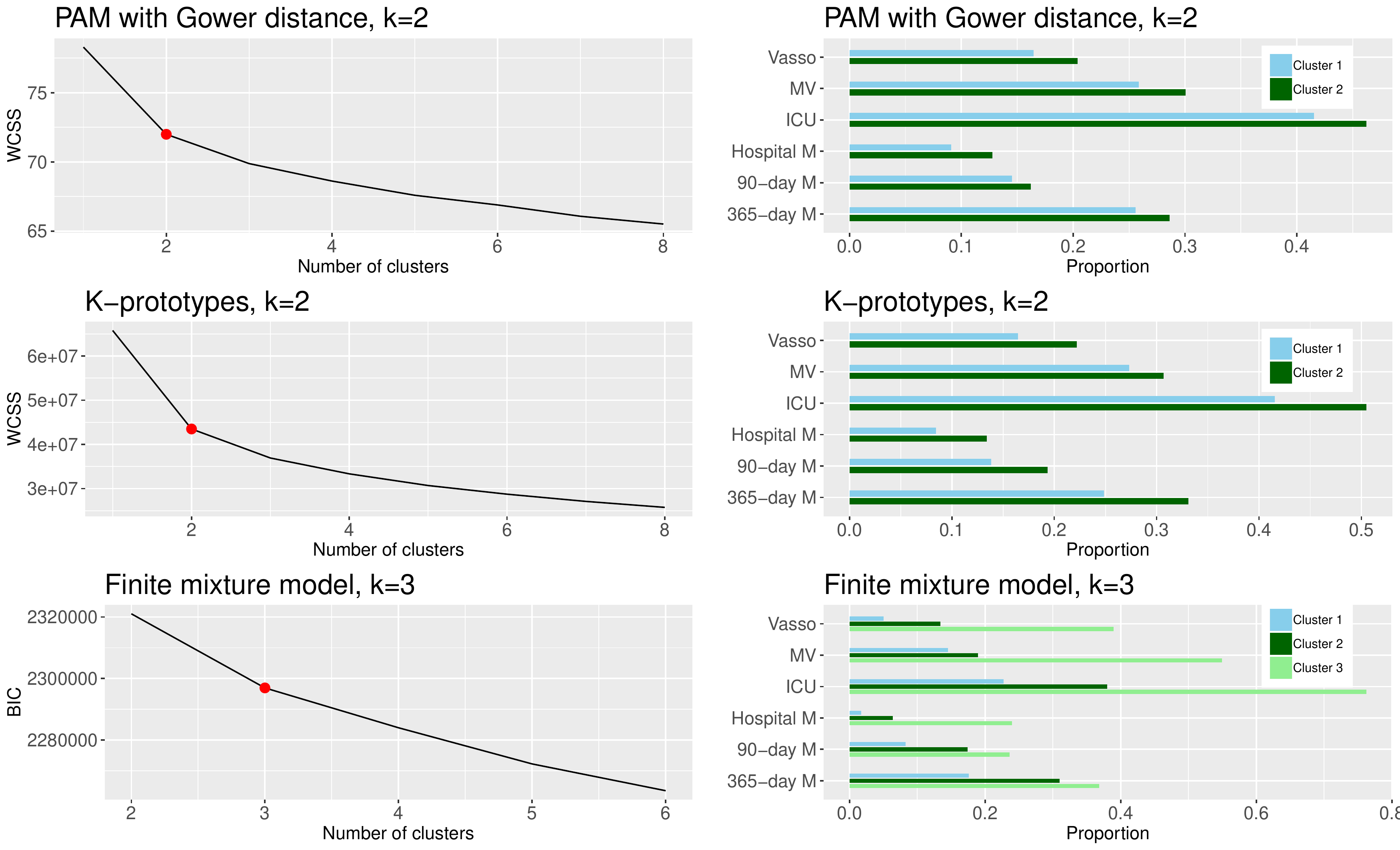}
\caption{Clinical endpoints across clusters with the optimal number of clusters}
\label{fig:senecak2}
\end{figure}

Next, we re-applied the existing methods by first selecting the method-specific optimal number of clusters. The number of clusters versus the total WCSS or BIC values are shown in left column of Figure~\ref{fig:senecak2}.
We found that the optimal number of clusters was 2 for PAM with Gower distance and for K-prototypes, and 3 for FMM. We once again observed that the clustering results were dominated by gender when using PAM with Gower distance. The proportion of men was 1.2\% in Cluster 1 and 98.8\% in Cluster 2. The two clusters were quite similar in terms of clinical endpoints. Similarly, the 2 clusters identified by K-prototypes were not quite distinct in terms of clinical endpoints as well. The FMM identified 3 clusters with quite different distribution of clinical endpoints, but the HyDaP algorithm was able to identify one more cluster with distinct clinical features.

\section{Discussion}
\label{sec:dis}

We proposed a hybrid density- and partition-based clustering (HyDaP) algorithm to conduct variable selection and identify clusters in data consisting of mixed types of variables. Our algorithm involves a pre-processing step to identify the data structure formed by continuous variables and to select important variables, and a clustering step to determine the cluster membership. In the clustering step, we proposed a dissimilarity measure that balances the contributions between continuous and categorical variables, which the existing clustering methods do not offer. Through simulation studies, we showed that the proposed HyDaP algorithm is robust to different data structures and outperforms or at par commonly used methods. We also defined 3 different data structures to help understand data and better interpret clustering results. Our method successfully identified four clinically meaningful sepsis phenotypes for data extracted from EHR of multiple health care systems. The resulting phenotypes are highly associated with several clinical endpoints.

Our HyDaP algorithm inherits the limitations of sparse K-means. For data under the \textit{natural cluster structure}, if the continuous variables contain many outliers or excessive zeros (a.k.a. zero-inflated), the sparse K-means procedure cannot correctly identify variables with high contributions. Another situation that could affect the later steps in our method and lead to unsatisfactory results is that when data contains continuous variables of the same value for the majority of subjects while other few have different values. We also suggest checking variables that have similar clinical meanings or highly clinically related before clustering and only keep one of them to avoid existence of within-cluster correlations.

Clustering has emerged as one of the essential and popular techniques for discovering patterns in data or disease phenotypes. Although clustering methods keep evolving to cope with increasing complexity in data, certain features in data sets could limit the utilization of the existing approaches. Unlike genetics or genomics data, data collected from clinical settings often include various types. Our proposed algorithm overcomes the drawbacks of the commonly used clustering algorithms therefore the results from using our method may be more helpful to clinicians in making good medical decisions.

\bibliographystyle{apalike} \bibliography{ref}

\clearpage

\appendix

\section{\\Variables used in SENECA data analysis}

Age\\
Gender: categorical variable; 2 levels (male/female)\\
Race: categorical variables; 3 levels (white/black/hispanic)\\
Maximum temperature within 6 hours of ER presentation\\
Maximum heart rate within 6 hours of ER presentation\\
Minimum systolic blood pressure within 6 hours of ER presentation\\
Maximum respiration rate within 6 hours of ER presentation\\
Maximum albumin within 6 hours of ER presentation\\
Maximum Cl within 6 hours of ER presentation\\
Maximum erythrocyte sedimentation rate (ESR) within 6 hours of ER presentation\\
Maximum hemoglobin within 6 hours of ER presentation\\
Maximum bicarbonate within 6 hours of ER presentation\\
Maximum Sodium within 6 hours of ER presentation\\
Minimum Glasgow Coma Scale (GCS) within 6 hours of ER presentation\\
Elixhauser Score\\
Maximum white blood cell within 6 hours of ER presentation\\
Maximum bands within 6 hours of ER presentation\\
Maximum creatinine within 6 hours of ER presentation\\
Maximum bilirubin within 6 hours of ER presentation\\
Maximum troponin within 6 hours of ER presentation\\
Maximum lactate within 6 hours of ER presentation\\
Maximum alanine aminotransferase (ALT) within 6 hours of ER presentation\\
Maximum aspartate aminotransferase (AST) within 6 hours of ER presentation\\
Maximum C-reactive protein within 6 hours of ER presentation\\
Maximum international normalized ratio (INR) within 6 hours of ER presentation\\
Maximum glucose within 6 hours of ER presentation\\
Maximum Platelets within 6 hours of ER presentation\\
Maximum blood urea nitrogen (BUN) within 6 hours of ER presentation\\
Oxygen saturation (SaO2)\\
Minimum partial pressure of oxygen (PaO2) within 6 hours of ER presentation

\end{document}